\documentclass{article} % For LaTeX2e
\usepackage{iclr2023_conference,times}

\usepackage{amsmath,amsfonts,bm}

\def\eqref#1{equation~\ref{#1}}
\def\1{\bm{1}}

\DeclareMathAlphabet{\mathsfit}{\encodingdefault}{\sfdefault}{m}{sl}
\SetMathAlphabet{\mathsfit}{bold}{\encodingdefault}{\sfdefault}{bx}{n}

\usepackage{hyperref}
\usepackage{url}
\usepackage{graphicx}
\usepackage{wrapfig,lipsum,booktabs}

\usepackage{amsmath, amssymb}
\usepackage{amsthm}
\theoremstyle{plain}
\newtheorem{theorem}{Theorem}[section]
\theoremstyle{definition}

\newtheorem{example}[theorem] {Example}

\usepackage{algorithm}
\usepackage{algpseudocode}
\algnewcommand{\Input}[1]{%
  \State \textbf{Input:}
  \Statex \hspace*{\algorithmicindent}\parbox[t]{.8\linewidth}{\raggedright #1}
}
\algnewcommand{\Initialize}[1]{%
  \State \textbf{Initialize:}
  \Statex \hspace*{\algorithmicindent}\parbox[t]{.8\linewidth}{\raggedright #1}
}

\title{Efficient, probabilistic analysis of \\combinatorial neural codes}

\author{Thomas F. Burns \\%\thanks{https://tfburns.com/, t.f.burns@gmail.com} \\
Neural Coding and Brain Computing Unit\\
OIST Graduate University, Okinawa, Japan \\
\texttt{thomas.burns@oist.jp} \\
\And
Irwansyah \\
Department of Mathematics \\
University of Mataram\\
West Nusa Tenggara, Indonesia \\
\texttt{irw@unram.ac.id} \\
}

\iclrfinalcopy % Uncomment for camera-ready version, but NOT for submission.
\begin{document}

\maketitle

\begin{abstract}
Artificial and biological neural networks (ANNs and BNNs) can encode inputs in the form of combinations of individual neurons' activities. These combinatorial neural codes present a computational challenge for direct and efficient analysis due to their high dimensionality and often large volumes of data. Here we improve the computational complexity -- from factorial to quadratic time -- of direct algebraic methods previously applied to small examples and apply them to large neural codes generated by experiments. These methods provide a novel and efficient way of probing algebraic, geometric, and topological characteristics of combinatorial neural codes and provide insights into how such characteristics are related to learning and experience in neural networks. We introduce a procedure to perform hypothesis testing on the intrinsic features of neural codes using information geometry. We then apply these methods to neural activities from an ANN for image classification and a BNN for 2D navigation to, without observing any inputs or outputs, estimate the structure and dimensionality of the stimulus or task space. Additionally, we demonstrate how an ANN varies its internal representations across network depth and during learning.
\end{abstract}

\section{Introduction}\label{sec:intro}

To understand the world around them, organisms' biological neural networks (BNNs) encode information about their environment in the dynamics of spikes varying over time and space. Artificial neural networks (ANNs) use similar principles, except instead of transmitting spikes they usually transmit a real-valued number in the range of $[0,1]$ and their dynamics are typically advanced in a step-wise, discrete manner. Both BNNs and ANNs adjust their internal structures, e.g., connection strengths between neurons, to improve their performance in learned tasks. This leads to encoding input data into internal representations, which they then transform into task-relevant outputs, e.g., motor commands. Combinatorial neural coding schemes, i.e., encoding information in the collective activity of neurons (also called `population coding'), is widespread in BNNs \citep{Averbeck2006, Osborne2008, Schneidman2011, Froudarakis2014, Bush2015, Stevens2018, Beyeler2019, Villafranca-Faus2021, Burns2022, Hannagane2021} and long-utilized in ANNs, e.g., in associative memory networks \citep{Little1974, Hopfield1984, Tsodyks1988, Adachi1997, Krotov2016}.

Advances in mathematical neuroscience \citep{Curto2008, Curto2019-1} has led to the development of analyses designed to understand the combinatorial properties of neural codes and their mapping to the stimulus space. Such analyses were initially inspired by the combinatorial coding seen in place cells \citep{Moser2008}, where neurons represent physical space in the form of ensemble and individual activity \citep{Brown2006, Fenton2008}. Place fields, the physical spatial areas encoded by place cells, can be arranged such that they span multiple spatial dimensions, e.g., 3D navigation space in bats \citep{Yartsev2013}. They can also encode for `social place' \citep{Omer2018}, the location of conspecifics. Just as these spatial and social dimensions of place (external stimuli) may be represented by combinatorial coding, so too may other dimensions in external stimuli, such as in vision \citep{Fujii1996, Panzeri2001, Averbeck2006, Froudarakis2014, Fetz2016}.

In place cells, the term receptive field (RF) or place field may intuitively be thought of as a physical place. In the context of vision, for example, we may think of RFs less spatially and more abstractly as representing stimuli features or dimensions along which neurons may respond more or less strongly, e.g., features such as orientation, spatial frequency, or motion \citep{Niell2008, Juavinett2015}. Two neurons which become activated simultaneously upon visual stimuli moving to the right of the visual field may be said to share the RF of general rightward motion, for example. We may also think of RFs even more abstractly as dimensions in general conceptual spaces, such as the reward--action space of a task \citep{Constantinescu2016}, visual attributes of characters or icons \citep{Aronov2017}, olfactory space \citep{Bao2019}, the relative positions people occupy in a social hierarchy \citep{Park2021}, and even cognition and behaviour more generally \citep{Bellmund2018}.

In the method described in \cite{Curto2019-1}, tools from algebra are used to extract the combinatorial structure of neural codes. The types of neural codes under study are sets of binary vectors $\mathcal{C}\subset\mathbb{F}_2^n$, where there are $n$ neurons in states $0$ (off) and $1$ (on). The ultimate structure of this method is the \textit{canonical form} of a neural code $CF(\mathcal{C})$. The canonical form may be analysed topologically, geometrically, and algebraically to infer features such as the potential convexity of the receptive fields (RFs) which gave rise to the code, or the minimum number of dimensions those RFs must span in real space. Such analyses are possible because $CF(\mathcal{C})$ captures the minimal essential set of combinatorial descriptions which describe all existing RF relationships implied by $\mathcal{C}$. RF relationships (whether and how RFs intersect or are contained by one-another in stimulus space) are considered to be implied by $\mathcal{C}$ by assuming that if two neurons become activated or spike simultaneously, they likely receive common external input in the form of common stimulus features or common RFs. Given sufficient exploration of the stimulus space, it is possible to infer topological features of the global stimulus space by only observing $\mathcal{C}$ \citep{Curto2008, Mulas2020}. To the best of our knowledge, these methods have only been developed and used for small examples of BNNs. Here we apply them to larger BNNs and to ANNs (by considering the co-activation of neurons during single stimulus trials).

Despite the power and broad applicability of these methods \citep{Curto2008, Curto2019-1, Mulas2020}, two major problems impede their usefulness: (1) the computational time complexity of the algorithms to generate $CF(\mathcal{C})$ is factorial in the number of codewords $\mathcal{O}(nm!)$\footnote{$n$ is the number of neurons and $m$ is the number of codewords. In most datasets of interest $n\ll m$.}, limiting their use in large, real-world datasets; and (2) there is no tolerance for noise in $\mathcal{C}$, nor consideration given towards the stochastic or probabilistic natures of neural firing. We address these problems by: (1) introducing a novel method for improving the time complexity to quadratic in the number of neurons $\mathcal{O}(n^2)$ by computing the generators of $CF(\mathcal{C})$ and using these to answer the same questions; and (2) using information geometry \citep{Nakahara2002, Amari2016} to perform hypothesis testing on the presence/absence of inferred geometric or topological properties of the stimulus or task space. As a proof of concept, we apply these new methods to data from a simulated BNN for spatial navigation and a simple ANN for visual classification, both of which may contain thousands of codewords.

\section{Preliminaries}\label{sec:prelims}
Before describing our own technical developments and improvements, we first outline some of the key mathematical concepts and objects which we use and expand upon in later sections. For more detailed information, we recommend referring to \cite{Curto2008, Curto2019-1}.

\subsection{Combinatorial neural codes}
Let $\mathbb{F}_2=\{0,1\}, [n]=\{1,2,\dots,n\},$ and $\mathbb{F}_2^n=\{a_1a_2\cdots a_n|a_i\in\mathbb{F}_2,\;\text{for all}\;i\}.$ A codeword is an element of $\mathbb{F}_2^n.$ For a given codeword $\mathbf{c}=c_1c_2\cdots c_n,$, we define its support as $supp(\mathbf{c})=\{i\in [n]|c_i\not=0\}$, which can be interpreted as the unique set of active neurons in a discrete time bin which correspond to that codeword. A combinatorial neural code, or a code, is a subset of $\mathbb{F}_2^n.$ The support of a code $C$ is defined as $supp(C)=\{S\subseteq [n]|S=supp(\mathbf{c})\;\text{for some}\;\mathbf{c}\in C\}$, which can be interpreted as all sets of active neurons represented by all corresponding codewords in $C$. % The support of a codeword is usually .

\noindent Let $\Delta$ be a subset of $2^{[n]}.$ The subset $\Delta$ is an {\it abstract simplicial complex} if for any $S\in \Delta,$ the condition $S'\subseteq S$ gives $S'\in \Delta,$ for any $S'\subseteq S.$  In other words, $\Delta\subseteq 2^{[n]}$ is an abstract simplicial complex if it is closed under inclusion. So, the simplicial complex for a code $C$ can be defined as 
\[\Delta(C)=\left\{S\subseteq [n]|S\subseteq supp(\mathbf{c}),\;\text{for some}\;\mathbf{c}\in C\right\}.\]
\noindent A set $S$ in a simplicial complex $\Delta$ is referred to as an $(|S|-1)$-simplex. For instance, a set with cardinality 1 is called $0$-simplex (geometrically, a point), a set with cardinality 2 is called a $1$-simplex (geometrically, an edge), and so on. Let $S$ be an $m$-simplex in $\Delta.$ Any $S'\subseteq S$ is called a face of $S.$

\subsection{Simplicial complexes and topology}
Let $C\subseteq \mathbb{F}_2^n$ be a code and $\Delta(C)$ be the corresponding simplicial complex of $C.$ From now on, we will use $\Delta$ to denote the corresponding simplicial complex of a code $C.$ Define $\Delta_m$ as a set of $m$-simplices in $\Delta.$ Define
\[C_m = \left\{\sum_{S\in \Delta_m}\alpha_S S\left|\right. \alpha_S\in\mathbb{F}_2,\forall S\in\Delta_m\right\}.\]
\noindent The set $C_m$ forms a vector space over $\mathbb{F}_2$ whose basis elements are all the $m$-simplicies in $\Delta_m.$ Now, define the chain complex $C_*(\Delta,\mathbb{F}_2)$ to be the sequence $\left\{C_m\right\}_{m\geq 0}.$ For any $m\geq 1,$ define a linear transformation $\partial_m: C_m\rightarrow C_{m-1},$ where for any $\sigma\in \Delta_m,$ $\partial_m(\sigma)=\sum_{i=0}^m\sigma^i,$ with $\sigma^i\in \Delta_{m-1}$ as a face of $\sigma,$ for all $i=0,\dots,m.$ Moreover, the map $\partial_m$ can be extended linearly to all elements in $C_m$ as follows
\[\partial_m\left(\sum_{S\in \Delta_m}\alpha_S S\right)=\sum_{S\in \Delta_m}\alpha_S \partial_m(S).\]
\noindent Define the $m$-th mod-2 homology group of $\Delta$ as
\[H_m(\Delta,\mathbb{F}_2)=\displaystyle{\frac{Ker\left(\partial_m\right)}{Im\left(\partial_{m+1}\right)}}\]
\noindent for all $m\geq 1$ and 
\[H_0(\Delta,\mathbb{F}_2)=\displaystyle{\frac{C_0}{Im\left(\partial_{1}\right)}}.\]
\noindent Note that $H_m(\Delta,\mathbb{F}_2)$ is also a vector space over $\mathbb{F}_2,$ for all $m\geq 0.$ So, the mod-2 $m$-th Betti number $\beta_m(\Delta)$ of a simplicial complex $\Delta$ is the dimension of $H_m(\Delta,\mathbb{F}_2).$ The $\beta_m(\Delta,\mathbb{F}_2)$ gives the number of $m$-dimensional holes in the geometric realisation of $\Delta.$

\subsection{Canonical form}
% Define and describe the added information in the canonical form. including the algorithm to construct it (basic description only -- of Algorithm 2 from \cite{Curto2019-1}).
Let $\sigma$ and $\tau$ be subsets of $[n],$ where $\sigma\cap\tau =\varnothing.$ The polynomial of the form $\prod_{i\in\sigma}x_i\prod_{j\in\tau}(1-x_j)\in\mathbb{F}_2[x_1m\dots,x_n]$ is called a {\it pseudo-monomial}. In a given ideal $\mathcal{J}\subseteq \mathbb{F}_2[x_1,\dots,x_n],$ a pseudo-monomial $f$ in $J$ is said to be minimal if there is no pseudo-monomial $g$ in $\mathcal{J}$ with $\deg(g)<\deg(f)$ such that $f=gh$ for some $h\in\mathbb{F}_2[x_1,\dots,x_n].$  For a given code $C\subseteq \mathbb{F}_2^n,$ we can define a {\it neural ideal} related to $C$ as $\mathcal{J}_C=\langle \rho_{\mathbf{c}'}|\mathbf{c}'\mathbb{F}_2^n-C\rangle,$ where $\rho_{\mathbf{c}'}$ is a pseudo-monomial of the form $\prod_{i\in supp\left(\mathbf{c}'\right)}x_i\prod_{j\not\in supp\left(\mathbf{c}'\right)}\left(1-x_j\right).$ A set of all minimal pseudo-monomials in $\mathcal{J}_C,$ denoted by $CF(\mathcal{J}_C)$ or simply $CF(C),$ is called the {\it canonical form} of $\mathcal{J}_C.$ Moreover, it can be shown that $\mathcal{J}_C=\langle CF(C)\rangle.$ Therefore, the canonical form $CF(C)$ gives a simple way to infer the RF relationships implied by all codewords in $C.$ One way to calculate the $CF(C)$ is by using a recursive algorithm described in \cite{Curto2019-1}. For a code $C=\{\mathbf{c}_1,\dots,\mathbf{c}_{|C|}\},$ the aforementioned algorithm works by constructing canonical forms $CF(\varnothing), CF\left(\{\mathbf{c}_1\}\right), CF\left(\{\mathbf{c}_1,\mathbf{c}_2\}\right),\dots,CF\left(C\right),$ respectively. In each stage, the algorithm evaluates polynomials, checks divisibility conditions, and adds or removes polynomials from a related canonical form.

\section{Methods}\label{sec:methods}
Our main methodological contributions are: (1) improving the computational complexity of the analyses relying on computing $CF(\mathcal{C})$ (see Algorithm \ref{alg:ours}); and (2) using information geometry to identify whether identified algebraic or topological features are statistically significant.

\subsection{Computing and analysing the canonical form's generators}
We may perform the same analyses as in \cite{Curto2019-1} in quadratic time by using Algorithm \ref{alg:ours} to construct the generators of $CF(\mathcal{C})$ rather than constructing $CF(\mathcal{C})$ itself (as in Algorithm 2 of \cite{Curto2019-1}). Illustrative of this efficiency, representative experimental data with $25$ neurons and $46$ codewords took $<1$ second to analyse on a high-end desktop PC (Intel i7 CPU and 64GB of memory), compared to 2 minutes 57 seconds using Algorithm 2 from \cite{Curto2019-1}.

\begin{algorithm}
\caption{Algorithm for computing generators of $CF(\mathcal{C})$}\label{alg:ours}
\begin{algorithmic}
%\Require $n \geq 0$
%\Ensure $y = x^n$
\Input{$M$ = $\mathcal{C}\subset\mathbb{F}_2^n$ as a $patterns \times neurons$ matrix}
\Initialize{
$D \gets \text{empty list}$ \Comment{Stores the monomials.}\\
\State $P \gets \text{empty list}$ \Comment{Stores the mixed monomial constructor tuples ($\sigma$,$\tau$).}\\
\State $B \gets \text{empty list}$ \Comment{Stores the mixed monomial constructor tuples ($\tau$,$\sigma$).}
}
\For{each column $i$ of $M$}
    \For{each column $j$ of $M$}
        \State $s \gets \sum_k (i \cdot j)_k$
        \If{$s<1$} \Comment{The pair $i,j$ have disjoint receptive fields.}
            \State $\text{ append } \{i,j\} \text{ to } D$
        \Else
        \State $j' \gets j-1$
        \State $b \gets \sum_{k=1} (i \cdot j')_k$
            \If{$b=0$} \Comment{The receptive field of $j$ is a subset of receptive field of $i$.}
                \State $\text{ append } (i,j) \text{ to } P$
                \State $\text{ append } (j,i) \text{ to } B$
            \EndIf
        \EndIf
    \EndFor
\EndFor
\end{algorithmic}
\end{algorithm}

% Output: Sets $D, P$, and $B$ which generate $\mathcal{J}_\mathcal{C}$, the superset of $CF(\mathcal{C})$.

% \noindent Step 2: For each column $i$ of $M$, calculate the sum of its dot product with every other column $j$ and set this as $s$. If $s<1$, add the set ${i,j}$ to $D$ (the pair $i,j$ have disjoint receptive fields).

% \noindent Step 3: For each column $i$ of $M$, set every other column $j$ equal to $j-1$. Then, calculate the sum $i \cdot j$ and set this as $b$. If $b=0$, add $(i,j)$ to $P$ and $(j,i)$ to $B$ ($j$ is a subset of $i$).

Generating desired elements of $\mathcal{J}_\mathcal{C}$ is then straightforward: monomials are supersets of disjoint pairs (from $D$) where each pair set has one element shared with at least one other disjoint pair set in the superset; mixed monomials are all possible combinations of first ($\sigma$ set) and second ($\tau$ set) elements in the tuples of $P$ (or vice-versa for $B$) -- we do not allocate all of these elements but instead store the set constructors; and the negative monomial appears if and only if the all 1s codeword exists (which involves a simple summing check on columns of $M$).

\subsection{Information geometry for combinatorial neural codes}

Let $N$ be the finite number of time bins for data of the neural activity patterns on $n$ neurons. For any $S\subseteq [n],$ let $v(S)\in\mathbb{F}_2^n,$ where $supp(v(S))=S$ and 
\begin{equation*}\displaystyle{\mathbf{P}_{v(S)}=\frac{\#\{v(S)\}}{N}}.\end{equation*}
We would like to find the parameters $\mathbf{\theta}=\left(\theta_{S_1},\theta_{S_2},\dots,\theta_{S_{2^n-1}}\right),$ where $S_i\subseteq [n],$ $S_i\not=\varnothing,$ and $S_{2^n-1}=[n],$ such that the following exponential function
\[\mathbf{P}(\mathbf{x},\mathbf{\theta})=exp\left(\sum_{S\subseteq [n],S\not=\varnothing}\theta_S x_S -\psi\right),\]
where $x_S=\prod_{i\in S}x_i$ and $\psi=-\log(P_{v(\varnothing)}),$ describes a neural activity pattern from the given neural activity data. 
We can calculate $\theta_S$ using the following formula for any $S\subseteq [n],$ where $S\not=\varnothing,$ 
\begin{equation*}\label{full}
        \theta_S = \displaystyle{\log\left(\frac{P_{v(S)}}{P_{v(\varnothing)}\prod_{S'\subsetneq S,S'\not=\varnothing}exp\left(\theta_{S'}\right)}\right),}
\end{equation*}
\begin{equation*}
    \eta_W=\sum_{S\subseteq [n], S\supseteq W}\mathbf{P}_{v(S)}.
\end{equation*}

Given a $\theta$-coordinate, we can calculate the associated $G(\theta)=\left(g_{A,B}^\theta\right)_{A,B\subseteq [n]}$ matrix using the following formula,
\begin{equation}\label{gab}
    \begin{array}{lll}
       g_{A,B}^\theta  & = & E_\theta\left(X_AX_B\right)-\eta_A\eta_B \\
       & & \\
         & = & \displaystyle{\sum_{W\supseteq A\cup B} e^{-\psi}\prod_{W'\subseteq W\atop W'\not=\varnothing}e^{\theta_{W'}}}-\eta_A\eta_B,\\
         & & \\
         %& & \\
         & = & \displaystyle{\sum_{W\supseteq A\cup B} e^{-\psi}e^{\sum_{W'\subseteq W\atop W'\not=\varnothing}\theta_{W'}}}-\eta_A\eta_B,
    \end{array}
\end{equation}
where $\psi = -\log(P_{v(\varnothing)}).$ 

\begin{example}
Let $n=4, A=\{1,2\},$ and $B=\{2,4\},$ then 
\[\begin{array}{lll}
g_{A,B}^\theta & = & \displaystyle{\sum_{W\supseteq \{1,2,4\}} e^{-\psi}\prod_{W'\subseteq W\atop W'\not=\varnothing}e^{\theta_{W'}}}\\
& & \\
& = & \displaystyle{e^{-\psi}\prod_{W'\subseteq \{1,2,4\}\atop W'\not=\varnothing}e^{\theta_{W'}}+e^{-\psi}\prod_{W'\subseteq \{1,2,3,4\}\atop W'\not=\varnothing}e^{\theta_{W'}}}\\
& &\\
 & = & \left(e^{\theta_{\{1\}}+\theta_{\{2\}}+\theta_{\{4\}}+\theta_{\{1,2\}}+\theta_{\{1,4\}}+\theta_{\{2,4\}}+\theta_{\{1,2,4\}}-\psi}\right.\\
 & & \left.+e^{\theta_{\{1\}}+\theta_{\{2\}}+\theta_{\{3\}}+\theta_{\{4\}}+\theta_{\{1,2\}}+\theta_{\{1,3\}}+\theta_{\{1,4\}}+\theta_{\{2,3\}}+\theta_{\{2,4\}}+\theta_{\{3,4\}}+\theta_{\{1,2,3\}}+\theta_{\{1,2,4\}}+\theta_{\{1,2,3,4\}}-\psi}\right)\\
 & & -\eta_{\{1,2\}}\eta_{\{2,4\}}
\end{array}\]
\end{example}

\subsection{Hypothesis testing for algebraic and topological features}

Using the previous sections, we can now perform hypothesis testing on specific RF relationships or topological features such as holes.

Given $\mathbf{P}_{v(S)}$ for all $S\subseteq [n]$ as in the previous subsection, we can calculate $\eta_W,$ for all $W\subseteq [n],$ where $\eta_W$ is equal to $E\left(\prod_{i\in W}x_i\right)=Prob\{x_i=1,\forall i\in W\},$ using the following formula.
\begin{equation*}
    \eta_W=\sum_{S\subseteq [n], S\supseteq W}\mathbf{P}_{v(S)}
\end{equation*}
Given a set of neurons $A\subseteq [n],$ where $|A|=k$, we want to test whether there is a $k$-th order interaction between neurons in $A$ or not. We can do this by hypothesis testing as follows.
\begin{enumerate}
    \item Calculate $\theta_S$ and $\eta_W,$ for all $S,W\subseteq [n].$
    
    \item Specify a coordinate for $\mathbf{P}(\mathbf{x};\mathbf{\eta},\mathbf{\theta})$ based on $A$ as
    \[\zeta_k^A=\left(\mathbf{\eta}_{k-}^A;\mathbf{\theta}_k^A\right),\]
    where $\mathbf{\eta}_{k-}^A=\left(\eta_H\right)_{H\subseteq [n],|H|\leq k}$ and $\mathbf{\theta}_k^A=\left(\theta_H\right)_{H\subseteq [n],|H|> k}.$
    
    \item Set the corresponding null hypothesis coordinate as
    \[\zeta_k^0=\left(\mathbf{\eta}_{k-}^A;\mathbf{\theta}_k^0\right),\]
    where $\eta_A=0,$ $\eta_H$ is equal to the previous step except for $H=A,$ and $\mathbf{\theta}_k^0$ is equal to the one in the previous step.
    
    \item Determine the corresponding $G(\theta)=\left(g_{A,B}^\theta\right)_{A,B\subseteq [n]}$ matrix related to $\theta$-coordinate using equation~\ref{gab}. Arrange the rows and columns of $G(\theta)$ such that 
    \[G(\theta)=\left(\begin{array}{cc}
    A_\theta & B_\theta\\
    B_\theta^T & D_\theta
    \end{array}\right),\]
    where $A_\theta$ is the submatrix of $G(\theta)$ with row and column indices from all $H\subseteq [n]$ with $|H|\leq k$ and $D_\theta$ is the submatrix of $G(\theta)$ with row and column indices from all $H\subseteq [n]$ with $|H|> k.$
    
    \item Determine the corresponding $G(\eta)=\left(g_{A,B}^\eta\right)_{A,B\subseteq [n]}$ matrix related to $\eta$-coordinate using the equation $G(\eta)=G(\theta)^{-1}.$  We can write $G(\eta)$ in the form 
    \[G(\eta)=\left(\begin{array}{cc}
    A_\eta & B_\eta\\
    B_\eta^T & D_\eta
    \end{array}\right),\]
    where $A_\eta$ is the submatrix of $G(\eta)$ with row and column indices from all $H\subseteq [n]$ with $|H|\leq k$ and $D_\eta$ is the submatrix of $G(\eta)$ with row and column indices from all $H\subseteq [n]$ with $|H|> k.$
    
    \item Determine the corresponding $G(\zeta_k^A)$ matrix related to the mixed coordinate $\zeta_k^A$ with
    \[G(\zeta_k^A)=\left(\begin{array}{cc}
    A_{\zeta_K^A} & \mathbf{O}\\
    \mathbf{O} & D_{\zeta_K^A}
    \end{array}\right),\]
    where $A_{\zeta_K^A}=A_\theta^{-1}$ and $D_{\zeta_K^A}=D_\eta^{-1}.$
    
    \item Calculate the test statistic as follows
    \[\begin{array}{ccc}
         \lambda & = & \displaystyle{2\sum_{i=1}^N\log\left(\frac{\mathbf{P}(\mathbf{x}_i;\eta_{k-}^A,\theta_k^0)}{\mathbf{P}(\mathbf{x}_i;\eta_{k-}^A,\theta_k^A)}\right)}\\
         && \\
         & \approx & 2N\tilde{E}\left(\displaystyle{\log\left(\frac{\mathbf{P}(\mathbf{x};\eta_{k-}^A,\theta_k^0)}{\mathbf{P}(\mathbf{x};\eta_{k-}^A,\theta_k^A)}\right)}\right)\\
         && \\
         & \approx & 2ND\left[\mathbf{P}(\mathbf{x};\eta_{k-}^A,\theta_k^0);\mathbf{P}(\mathbf{x};\eta_{k-}^A,\theta_k^A)\right]\\
         && \\
         & \approx & \displaystyle{N g_{AA}^\zeta(\eta_A^0-\eta_A)}\\
         & & \\
         & \approx & \displaystyle{N g_{AA}^{\zeta_k^A} (\eta_A^0-\eta_A)}
    \end{array}\]
    where $g_{AA}^{\zeta_k^A}$ is the entry of the $G(\zeta_k^A)$ matrix.
    
   % \[g_{SS'}^\zeta(\theta_k^0)=E\left(x_Sx_{S'}\right)-\eta_S\eta_{S'},\]
    % with
    % \[x_S=\prod_{i\in S}x_i,\]
    % and
    % \[E\left(x_Sx_{S'}\right)=\sum_{\mathbf{x}\in\mathbb{F}_2^n}x_Sx_{S'}\mathbf{P}(\mathbf{x};\eta_{k-}^A,\theta_k^0)=P_{v(S\cup S')}.\]
    % E is expectation, defined differently for xsxs' and the log term
    % D is divergence
    
    \item Fix a level of significance $\alpha$ and find the value $\chi^2_{\alpha}(1)$ (chi-square value with significance level $\alpha$ and degree of freedom $1$) from the $\chi^2$ look-up table.
    
    \item Compare $\lambda$ and $\chi^2=\max\{\chi^2_\alpha(1),1-\chi^2_\alpha(1)\}$
    
    \begin{itemize}
        \item If $\lambda\geq \chi^2,$ there is a significant interaction between neurons in $A$ (reject the null hypothesis)
        \item Otherwise, there is no significant interaction between neurons in $A$ (accept the null hypothesis)
    \end{itemize}
\end{enumerate}

Since $G$ scales at $2^{n}$, we use a subset $M$ of all neurons, where $A \subset M$ and $|M|=10$. We pick a set $A$ relevant to the feature we want to test the significance of, and choose random neurons (without replacement) not already in $A$ for the remaining elements of $M$, repeating the test until we exhaust all neurons. We then correct for multiple comparisons and use $\alpha=0.05$ to detect whether there is a significant interaction in $A$.

The choice of $A$ depends on which feature we wish to analyse. When analysing whether or not two neurons are disjoint or not in their RFs (a monomial relationship in $CF(\mathcal{C})$), we set $A$ as those two neurons. When analysing whether the RF of $i$ is contained within the RF of $j$ (a mixed monomial relationship in $CF(\mathcal{C})$), we first set $A$ as those two neurons, and then set $A$ as $i$ with a random set of neurons (repeating this at least $5$ times, with different random sets, and correcting for multiple comparisons). For every dimension $m$ where $\beta_m(\Delta,\mathbb{F}_2)>0$, when analysing whether a hole is significant, we test for all possible sets $A \subset M \subset [n]$ which close the hole. If any test closes the hole, there is no hole, whereas if no test closes the whole, there is a hole.

\section{Applications}\label{sec:applications}

% \subsection{Computing generators of the canonical form}

% Explain in simple language (put more technical details in the methods section).

% \subsection{Hypothesis testing}

% Explain in simple language (put more technical details in the methods section).

% \subsection{Datasets}

% Given the generality of these methods, we decided to test both ANNs and BNNs. Experimental advances in \textit{in-vivo} 2-photon fluorescence imaging \cite{Kalatsky2003,Chen2013,Goldey2014} and electrophysiology \cite{} now allows researchers to observe the simultaneous activity of tens, hundreds, and even thousands of neurons over time in awake, intact animals. Combined with systematic variation of sensory stimuli or tasks, these studies offer us the opportunity to study whether combinatorial neural codes can infer the topological properties of the stimulus or task space. We chose to focus on spatial navigation and visual perception because \dots

% Unlike the experimental limitations of studying BNNs, researchers have complete access to the internal states of neurons in ANNs. ANNs can also be trained on tasks which emphasise the utility of this method.

\subsection{Spatial navigation in BNNs}
\begin{figure}[h]
    \centering
    \includegraphics[width=0.86\textwidth]{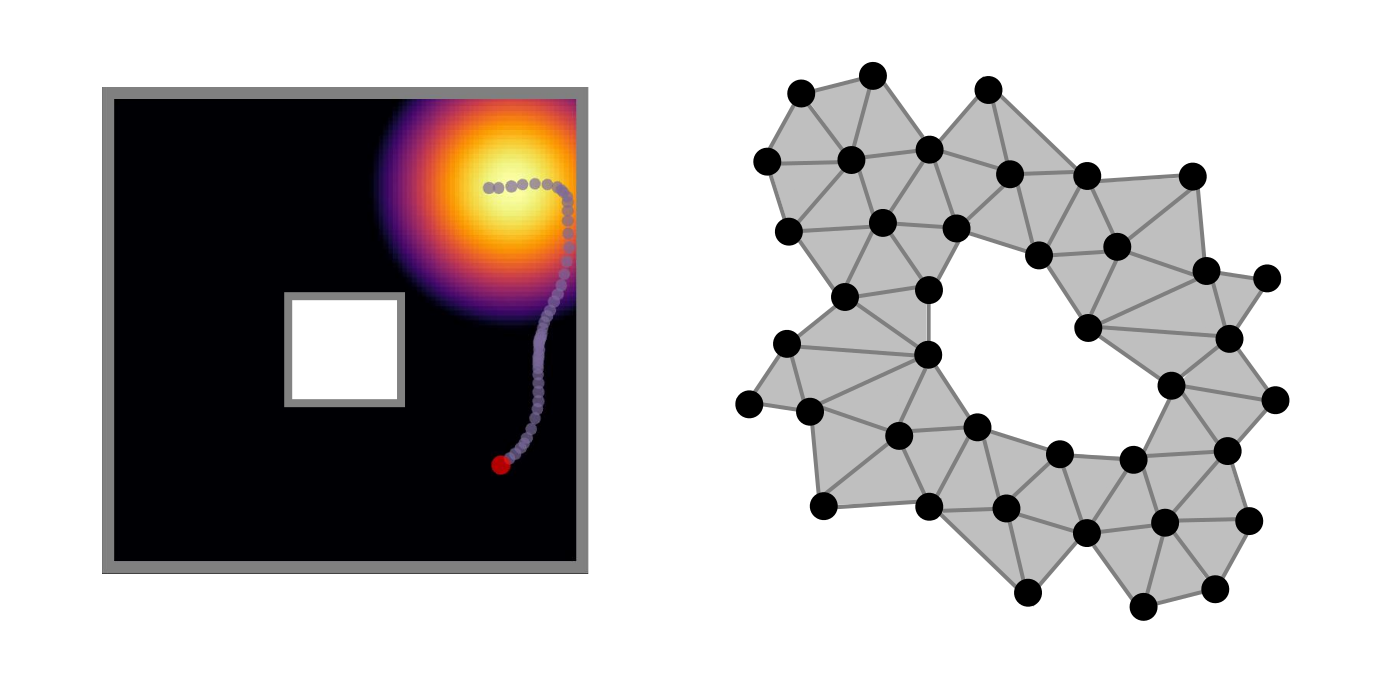}
    \caption{Example of a single place cell's receptive field and part of a simulated trajectory for an environment with $\beta_1=1$ (left) and a geometric realisation of $\Delta(\mathcal{C})$ constructed after a 30 minute random walk in that environment (right).}
    \label{fig:navigation}
\end{figure}

Using the RatInABox simulation package \citep{George2022}, we created simple 2D navigation environments with 0, 1, 2, or 3 holes in the first dimension. We used a random cover of 40 place cells modelled using Gaussians for the probability of firing and geodesic receptive field geometries. Starting at a random position, we then simulated random walks governed by Ornstein-Uhlenbeck processes for 30m, with parameters based on rat locomotion data in \citep{Sargolini2006}. We constructed a combinatorial neural code $\mathcal{C}$ using a window size of 10ms, allowing for up to 3,000 unique codewords. We constructed $\Delta(\mathcal{C})$ up to dimension 2 and calculated $\beta_1(\Delta,\mathbb{F}_2)$, with the hypothesis that $\beta_1$ would be equal to the respective number of holes in the environment. Figure \ref{fig:navigation} shows an example of a single place cell and part of a simulated trajectory for an environment with $\beta_1=1$ and a geometric realisation of $\Delta(\mathcal{C})$ constructed after a 30 minute random walk. Table \ref{tab:navigation-topology} shows the number of statistically significant holes found after different durations of the trajectories for environments with different topologies. Although after 10 minutes of a random walk some holes were occasionally detected, in all cases after 20 minutes all holes in the environment were detected consistently. There were a large number of of monomials across all conditions (all simulations had $>1000$) due to the covering nature of the RF arrangements. There were also a small number (all simulations had $<5$) of mixed monomials (RFs found to be subsets, significantly so, of other RFs).

%\begin{wraptable}{r}{7cm}
\begin{table}[h]
\begin{center}
\caption{Number of statistically significant holes found in 2D environments with 0, 1, 2, or 3 holes after 10, 20, or 30 minutes of simulated time. Quoted are the means $\pm$ S.D.s across 10 simulations, each with different random place cell coverings and trajectories. All simulations had $\beta_0=1$.}
\begin{tabular}{ccccc}
     & \multicolumn{4}{c}{\textit{Number of holes in the 2D environment}} \\
\textit{Time} & \textit{0}      & \textit{1}      & \textit{2}     & \textit{3}     \\ \hline
10 minutes  & $0 \pm 0$     & $0 \pm 0$     & $0.8 \pm 0.4$    & $1.3 \pm 0.46$     \\
20 minutes  & $0 \pm 0$     & $1 \pm 0$     & $2 \pm 0$    & $3 \pm 0$    \\
30 minutes  & $0 \pm 0$     & $1 \pm 0$     & $2 \pm 0$    & $3 \pm 0$   
\end{tabular}
\label{tab:navigation-topology}
\end{center}
\end{table}
%\end{wraptable}

% \subsection{Object manipulation in ANNs}

% Robotic arm navigating in 3D space to perform grasping tasks where there are ``off-limit'' areas (for safety reasons, let's say)

% \subsection{Optimal trajectories in 2D navigation}

% How many steps are required to get the same confidence of the homology/topology/algebra?

% Test numerically whether different types of walks can successfully detect homology of rooms or not (like https://journals.plos.org/ploscompbiol/article?id=10.1371/journal.pcbi.1000205)
% - random walks (Levy flight, random walks with different temperatures)
% - biased random walks (biased towards regions/quadrants of the room not yet explored)
% - maze completing algorithms

% To test:
% 1. simulate all walks for the same number of steps (S, S is large)
% 2. check the homology/algebraic relationships of and the related statistical confidence at different timepoints (different subsets of the data)
% 3. compare features \& confidence with group truth
% 4. discuss why some walks are closer to truth than others

\subsection{Visual classification in ANNs}

\begin{figure}[t]
    \centering
    \includegraphics[width=0.8\textwidth]{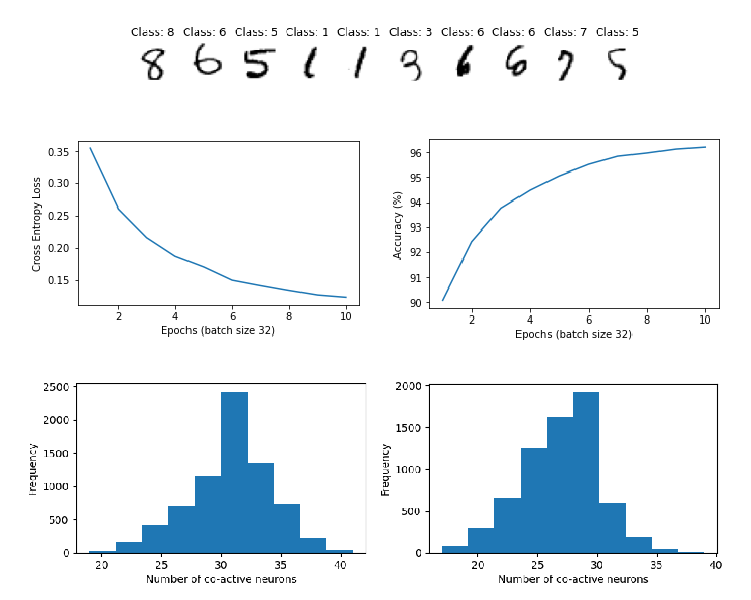}
    \caption{Top: Example MNIST digits and their classes. Middle: Test loss and accuracy over training epochs. Bottom: Histograms of the number of co-active neurons in each codeword, i.e., $|supp(\textbf{c})|$, for $\mathcal{C}_2$ after 1 (left) and 10 (right) training epochs.}
    \label{fig:mnist}
\end{figure}

We trained a multi-layer perceptron (MLP) to classify handwritten digits from the MNIST dataset \citep{lecun-mnisthandwrittendigit-2010} (see Figure \ref{fig:mnist}, top, for examples). The model consisted of an input layer with 784 neurons (the digit pixel values), followed by two hidden layers, each with 50 neurons using the rectified linear unit activation function and 20\% dropout. The final output layer consisted of 10 neurons (corresponding to the 10 digit class labels) and used a softmax activation function. The data was split into 50,000 digits for training, 10,000 for validation, and 10,000 for testing, allowing for up to 10,000 unique codewords in our analysis. The network was trained over 10 epochs with a batch size of 32 samples. The optimiser was stochastic gradient descent (with learning rate 0.01 and momentum 0.5) and the criterion was the cross-entropy loss between the one-hot vector of the true class labels and the output layer's activation for each sample. The MLP achieved $>96\%$ accuracy after 10 epochs (Figure \ref{fig:mnist}, middle).

After each epoch, test samples which the network did not see during training were fed through the network and the activity of all neurons in both hidden layers was recorded. The recorded activities for each hidden layer corresponding to each sample were then binarized about their means (calculated over all samples) to create a code $\mathcal{C}$ of size $10,000 \times 50$ for each layer, which we denote $\mathcal{C}_1$ for layer one and $\mathcal{C}_2$ for layer two.

\begin{table}[h]
\begin{center}
\caption{Algebraic and geometric properties of RFs of the codes $\mathcal{C}_1$ and $\mathcal{C}_2$ from the first and second hidden layers, respectively, of the MLP trained on MNIST classification. Numbers indicate counts of polynomials of that type. Monomials are reported as a tuple over orders (first, second, ..., n-th), e.g., `(5,3)' means there were 8 monomials, 5 of order 1 and 3 of order 2.}
\begin{tabular}{ccccccc}
\textit{Epoch}          & \multicolumn{2}{c}{\textit{1}} & \multicolumn{2}{c}{\textit{5}} & \multicolumn{2}{c}{\textit{10}} \\
\textit{Layer code}     & \textit{$\mathcal{C}_1$}    & \textit{$\mathcal{C}_2$}   & \textit{$\mathcal{C}_1$}    & \textit{$\mathcal{C}_2$}   & \textit{$\mathcal{C}_1$}    & \textit{$\mathcal{C}_2$}    \\ \hline
Monomials               & (25,9,1)       & (12,7)        & (5,1)          & (2)           & (1)            & -              \\
Mixed monomials         & 559,171        & 266,322       & 2,530          & 66            & 10             & 3              \\
Lower bound of dimension     & 4              & 3             & 3              & 2             & 2              & 1              \\
Negative monomials      & 0              & 0             & 0              & 0             & 0              & 0              \\
Local obstruction to convexity?      & No             & No            & No             & No            & No             & No             \\
Intersection complete? & No             & No            & No             & No            & No             & No            
\end{tabular}
\label{tab:mnist-algeom}
\end{center}
\end{table}

The codes $\mathcal{C}_1$ and $\mathcal{C}_2$ showed differences in their algebraic and geometric structures across training epochs, and also differed between themselves (Table \ref{tab:mnist-algeom}). In general, $\mathcal{C}_1$ had more overlapping RFs and spanned a larger number of real dimensions (assuming convexity) than $\mathcal{C}_2$. However, during training, we find both codes lower their dimensionality and gradually spread out their RFs to cover more of the space. This is also shown by the leftward shift between epoch 1 and 10 in the histograms of the number of co-active neurons in $\mathcal{C}_2$ (Figure \ref{fig:mnist}, bottom).

\section{Discussion}\label{sec:discussion}
We have shown it is possible to analyse the intrinsic geometry and topology of combinatorial neural codes from biological and artificial networks in an efficient and probabilistic manner. With these improved methods, we can now comfortably study codes with tens and even hundreds of thousands of codewords. We have shown how these methods can be used to better understand (with some statistically surety) how the internal representations of external inputs within these networks can change through learning, experience, and network depth.

Neuroscientists have shown combinatorial neural codes can occupy low-dimensional subspaces called \textit{neural manifolds} in the covariance of their neural activities \citep{Gallego2017, Feulner2021}. Trajectories and regions in these subspaces can correspond to task cognition, perceptual classification, and movement \citep{Cohen2020, Chung2021}. For example, \cite{Gardner2022} show the activity of populations of hundreds of grid cells within single modules of medial entorhinal cortex (a brain area partly responsible for navigation) occupy positions on a toroidal manifold. Positions on this manifold correspond to positions in the 2D space which the animal is navigating in.

These findings might lead us to believe combinatorial neural codes are intrinsically low-dimensional despite being embedded in the high-dimensional combinatorial space of neural activity. However, theoretical \citep{Bartolo2020} and experimental \citep{Rigotti2013} studies have shown that the dimensionality of these neural manifolds is influenced and often directly corresponds to the dimensionality of the task or learning under study. Indeed, the low-dimensional embeddings found by \cite{Gardner2022} are predicted by the two-dimensionality of the navigation (the underlying cause of the neural activity). Mathematically-optimal combinatorial neural codes and their RFs are also related to the dimensionality of the inputs those codes are attempting to represent \citep{Wang2013}. In more naturalistic and complex tasks, maintaining high-dimensional representations in the neural code may allow for increased expressibility but lower generalisability, whereas reducing to low-dimensional representations may allow for less expressibility but higher generalisability \citep{Fusi2016, Badre2021}. High-dimensional codes are often found in recordings from BNNs and are are often found when individual neurons encode for multiple input features, allowing linear read-out of a large number of complex or simple features \citep{Fusi2016}. Such neurons, for example in macaque inferotemporal cortex \citep{Higgins2021}, can also encode for very specific and independent higher-dimensional features.

This implies combinatorial neural codes can include mixtures of coding strategies which are simultaneously low- and high-dimensional. One of the key advantages of the techniques developed and applied in this study is that we can consider these different dimensionalities of coding at the same time. We don't reduce the embedding dimensionality to perform our analysis (which would be equivalent to assuming a low-dimensional code). We also don't try to map individual neuron responses to experimenter-known but network-unknown external, high-dimensional variables (which would be equivalent to assuming a high-dimensional code). Instead, we keep the full, original dimensionality of the data and can identify low- or high-dimensional features and response relationships at local and global levels simultaneously, all without reference to information external to the neural network. We also provide a method for testing the statistical significance of these features and relationships, again while maintaining the original high embedding dimension of the data. This allows us to avoid making any strong assumptions about dimensionality of the task, stimuli, or the corresponding neural code -- instead, we let the data speak for themselves.
% Also: https://twitter.com/madsarv/status/1494354955989307396
% https://www.biorxiv.org/content/10.1101/2020.12.29.424235v1
% https://journals.plos.org/ploscompbiol/article?id=10.1371/journal.pcbi.1008146

We do carry over some limitations from prior work, most prominently: (a) we assume joint-activity of neurons corresponds to common inputs or selectivity thereof; and (b) we binarize neural signals into `on' and `off' states. We suggest future work now focus on mitigating these limitations by: (a) performing causal inference tests on neural co-activations; and (b) considering polynomials over larger finite fields, e.g., $\mathbb{F}_4$, or extending these methods to more `continuous' structures, e.g., manifolds.

% Granted, we do carry over some limitations from prior work \cite{Curto2008, Curto2019-1}, most prominently: (a) we assume joint-activity of neurons corresponds to common inputs or selectivity thereof; and (b) we binarize neural signals into ``on'' and ``off'' states. Limitation (a) seems worse for BNNs than ANNs -- ANNs can have fully-observable structures and simultaneous inputs, whereas BNNs can have unknown structures and inputs, contain responses occurring at different transmission and processing speeds, and experimental limitations mean we can rarely track the activity of every relevant neuron. Conversely, limitation (b) seems worse for ANNs than BNNs -- BNNs mostly use binary signals (spikes) to transmit messages whereas ANNs use continuous values (which here we must binarize). Despite these limitations, we have shown reliable information can be found in 
% we could also test overlaps in a chain of RFs or edges in the simplicial complex of the code, etc., to test if C is connected
% test minimum spanning tree of the 1-skeleton of the simplicial complex
% We thank Tomoki Fukai, Emtiyaz Khan, Hiroyuki Nakahara, and members of the Neural Coding and Brain Computing Unit (OIST) for helpful discussions.

\bibliography{main}
\bibliographystyle{iclr2023_conference}

% \appendix
% \section{Appendix}

\end{document}